%% file: main.tex
\documentclass[letterpaper]{article} 
\usepackage{aaai2027}  
\usepackage[hyphens]{url}  
\usepackage{graphicx} 
\usepackage{natbib}  
\usepackage{caption} 
\usepackage{algorithm}
\usepackage{algorithmic}
\usepackage{amsmath}
\usepackage{subcaption}
\usepackage{pgfplots}
\usepackage{datatool}
\usepackage{appendix}
\usepgfplotslibrary{fillbetween}
\usepackage{newfloat}
\usepackage{listings}
\DeclareCaptionStyle{ruled}{labelfont=normalfont,labelsep=colon,strut=off} 
\floatstyle{ruled}
\newfloat{listing}{tb}{lst}{}
\floatname{listing}{Listing}

\usepackage{booktabs}

\title{SPADE-KD:\\Peer-to-Peer Collaborative Learning under Heterogeneous Settings}
\author {
    Harikrishna Kuttivelil\textsuperscript{\rm 1} \&
    Katia Obraczka\textsuperscript{\rm 2}
}
\affiliations {
    University of California, Santa Cruz\\
    \textsuperscript{\rm 1}hkuttive@ucsc.edu, \textsuperscript{\rm 2}obraczka@ucsc.edu
}

\begin{document}

\maketitle

\begin{abstract}
Fully decentralized, peer-to-peer collaborative learning is an emerging distributed deep learning paradigm that, unlike its centralized counterparts, eliminates the need for a central aggregator and disseminator, avoiding the single-point-of-failure and bottleneck issues they can bring.
Many such approaches -- including decentralized federated and gossip learning -- use \textit{parameter averaging}, requiring clients to have homogeneous model architectures, whereas others use knowledge distillation as an alternative while leaning on shared public data or centralized control infrastructure.
Most decentralized learning approaches also make a \textit{one-size-fits-all} assumption, i.e., they collaboratively train clients towards a single, globally optimal solution for all clients, when in reality, clients' underlying data distributions might differ such that a globally optimal solution may not be locally optimal for all clients.
To enable collaborative learning in environments where heterogeneity is present in both model architecture and in underlying data distribution, we present SPADE-KD, a fully decentralized serverless model distillation framework that optimizes collaboration towards personalized rather than global performance.
SPADE-KD leverages two mechanisms -- \textit{adaptive distillation} and \textit{utility-driven peer selection}.
Adaptive distillation modulates the strength of each incoming model transfer by the performance advantage of the incoming model compared to the native model measured by their task losses, such that a client will distill more aggressively from incoming models that outperform it on its own local data.
Utility-driven peer selection uses the loss-informed utility derived from adaptive distillation to rank prospective partners, focusing on exchanges that measurably improve the local model while suppressing low-utility transfers.
Neither mechanism relies on any centralized infrastructure or public datasets.
Using federated classification tasks, we demonstrate that SPADE-KD consistent outperforms naive decentralized distillation with lower loss and faster convergence and we provide ablation studies of the combined and individual effectivenesses of SPADE-KD's component mechanisms.
\end{abstract}


\input{sections/1_introduction}
\input{sections/2_background}
\input{sections/3_methods}
\input{sections/4_experiment}
\input{sections/5_conclusion}

\bibliography{references}
\input{appendix/6_appendix}


\end{document}

%% file: sections/1_introduction.tex
\section{Introduction}
As more data is generated at the network edge and utilized to train deep learning models, distributed model training over decentralized data has become an attractive structural paradigm for deep learning, mitigating privacy concerns and providing more autonomy for participating clients and improving communication efficiency between servers and/or clients.
While many centralized and semi-centralized distributed learning approaches are facilitated or coordinated through centralized servers, raising concerns about servers being singe points of failures or bottlenecks in collaborative learning among networks of clients, and motivating fully decentralized collaborative learning paradigms.

However, distributed learning -- centralized and decentralized -- must contend with issues emerging from the heterogeneity that is especially characteristic in realistic peer-to-peer settings.
Heterogeneity within underlying data distributions poses challenges to distributed learning approaches that target consensus towards a global solution, as they may not adequately serve networks of clients whose data distributions show incongruence enough that a globally optimal solution may not be the best solution -- or even an adequate one -- for all clients.
Heterogeneity in model quality as a result of training capacities or quantity and/or quality of clients' local data poses opportunities to bridge model performance disparity through collaboration but also challenges in preserving performance higher-quality models, mitigating training volatility, and convergence of local models.
While many federated learning approaches have been proposed to address these two axes of heterogeneity, heterogeneity in model architectures make it impossible to use the model parameter-averaging mechanisms at the core of many of them, and therefore knowledge must be instead transferred functionally -- via knowledge distillation.

Yet the scarce but growing landscape of existing distributed distillation methods inherit two assumptions that can fail in heterogeneous settings: (1) that every model has something to offer that is worth distilling from, and (2) that every potential peer client is worth collaborating with.
When peers hold heterogeneous data and unequal competence, both assumptions break — and a naive scheme can drag a capable agent down toward a weaker peer.
Even distribution distillation methods that confront that challenge these assumptions assume a level of heterogeneity low enough that distribution variance among clients' data can be overcome with universal collaboration among clients.
In the face of more challenging heterogeneity, we must consider both how a client should distill from models shared with it from other clients and how a client should select peers it wants to receive models from.

Our key insight is that both problems share a solution -- a client can measure the potential utility of a shared model on its own data and use this signal to determine how much to distill from each shared model and which clients it should query next --- i.e., one utility can govern both what a client learns and who it should learn from, in the context of collaborative learning.
This insight yields our proposed approach, Serverless Peer-to-Peer Adaptive Decentralized Knowledge Distillation (SPADE-KD), a framework that builds both the distillation mechanism and the peer selection mechanism from a single utility signal.
In SPADE-KD, each client computes an adaptive coefficient from the relative task loss of a peer's model versus its own native model, and uses it to modulate that peer's contribution in the distillation objective, promoting prospective teacher models that outperform its native model locally and surpressing those that do not.
Because this coefficient falls out of distillation as a by-product, we reuse it as a stored utility score that drives peer selection, with a neutral default for never-seen peers that yields exploration at no additional cost. Our contributions are:
\begin{itemize}
    \item A decentralized adaptive knowledge distillation method that derives a utility signal $\gamma$ from the comparative loss advantage of an incoming prospective teacher model and the native student model (Eq.~\ref{eq:akd-coefficient}) and applies it as a co-effecient to modulate distillation.
    \item The re-utilization of that same utility signal $\gamma$ as an indication of collaboration potential in the formulation of a multi-armed bandit problem for each client to select the optimal peers to request prospective teacher models from to best collaboratively train its native model and minimize loss over its local data.
    \item SPADE-KD, a fully decentralized collaborative deep learning framework, that utilizes decentralized adaptive knowledge distillation and utility-based peer selection to account for for networks of clients with heterogeneity in data distributions, model architectures, and model qualities within the network.
\end{itemize}

%% file: sections/2_background.tex
\section{Background \& Related Work}
Our work builds on the emerging domain of distributed, collaborative training among decentralized, autonomous clients with private data.
In the traditional deep learning paradigm, clients would be required to send some or all their data to a server for the server to then train the model, presenting issues with communication bandwidth and data privacy.

\paragraph{Federated \& Decentralized Learning}
Federated learning (FL)~\cite{og-fl}
proposed having a
central server disseminate copies of its model to clients and perform \textit{parameter averaging} of the returned client-trained model updates to periodically update its global model, making it unnecessary for clients to send their data directly to the server and providing consequent benefits in privacy and communication bandwidth utilization.
Personalized variations on FL have followed, e.g., proposing adaptations to its structure of collaboration and aggregation~\cite{personalized-fl} and using the cosine similarity of model progressions to produce differentiated server models trained by clusters of clients~\cite{cfl}, among other approaches.
To address the single point of failure and performance bottleneck issues raised by centralized approaches, decentralized FL approaches~\cite{dfl} have been proposed and proven competitive to centralized FL, including asynchronous, peer-to-peer gossip learning~\cite{gossip}.
These decentralized adaptations include some approaches that address data and training heterogeneity among clients, proposing mechanisms such as optimal peer selection and weighted aggregation to overcome heterogeneity and reach consensus on a global optimal solution~\cite{gossipfl}.
Other approaches target more localized optimal solutions by exploiting observable heterogeneity as clients train~\cite{dpfl}.

\paragraph{Distributed Knowledge Distillation}
The utilization of knowledge distillation in distributed, collaborative deep learning aims to overcome some of the detriments of parameter averaging, namely the requirement for homogeneous model architectures and the catastrophic forgetting incurred by the averaging of divergent client models.
Some decentralized FL approaches leverage knowledge distillation as an intermediary mechanism to discern compatible collaboration partners~\cite{pdfl-kd}.
Other proposals implement mutual distillation to optimize information exchange from peer-aggregated models and peer selection~\cite{dflstar} in networks of high data heterogeneity, supplementing but not supplanting the parameter averaging-based aggregation mechanism.
FL has also evolved to include federated knowledge distillation approaches~\cite{fedkd, proxyfl} that leverage mutual knowledge distillation to produce compact proxy models that can be sent with better communication efficiency over the network to then be aggregated via parameter averaging for better training robustness.
These approaches yield the additional benefit that because native client models are distilling into these compact proxy models and only these proxy models are being exchanged, only the proxy models must have identical model architectures, enabling the native client models to have heterogeneous architectures.
To enable collaborative decentralized learning between heterogeneous model architectures towards a globally optimal solution space, decentralized federated mutual learning~\cite{dfml} has been proposed, in which clients in a decentralized network take turns mutually distilling partners' models alongside their own over their local data, oscillating between local training and distillation.
However, this and similar approaches do not consider the possibility of data distributions being heterogeneous to the point of incongruence in which training jointly on clients' data -- directly or indirectly via client-trained model updates -- can be detrimental to the training of any single client.
To that end, one LLM-specific orchestrated gossip learning via distillation approach~\cite{knexafl} proposes to have a centralized matchmaker that learns throughout the collaborative training phase to form client pairs based on training performance and client-provided profiles.
To our knowledge, our approach is the first to provide a general solution to fully decentralized collaborative learning via knowledge distillation among clients that addresses both heterogeneous model architecture as well as data heterogeneity and training disparity.

%% file: sections/3_methods.tex
\section{Peer-to-Peer Adaptive Knowledge Distillation}

To facilitate collaboration in decentralized networks with heterogeneous data, models, and training history, we introduce Serverless Peer-to-Peer Adaptive Decentralized Knowledge Distillation -- SPADE-KD -- a fully decentralized approach that adaptively modulates distillation from received models using a utility signal derived from local task loss advantage and reuses that same signal to drive utility-based peer selection.

\paragraph{System Model}
We consider a fully decentralized network $\mathbf{N}$ of $n$ clients that collaboratively train without a central coordinator or a shared global model.
We model the network as an undirected small-world graph $G_{\mathbf{N}} = (V, E)$, where each vertex $i \in V$ (with $|V| = n$) is a client and each edge $(i, k) \in E$ is a logical communication link between two clients.
The set of candidate peers $M_i$ available to client $i$ is defined by reachability $M_i = \{\, k \in V : (i, k) \in E \,\}$ or provided by some peer sampling service $\Omega$ such that $M_i \sim \Omega(\mathbf{N})$.
In each collaboration round client $i$ may request models from a subset $\mathcal{C}_i \subseteq M_i$ of $m_i = |\mathcal{C}_i| \le |M_i|$ of these peers.
Each client $i$ holds a private local dataset $D_i$ and a native model $\theta_i$.
We consider the cases in which the network is heterogeneous, specifically along three axes:
(1) \emph{data heterogeneity} -- clients' local data $D_i$ may come from non-identical, potentially incongruent distributions;
(2) \emph{model heterogeneity} -- clients' native models $\theta_i$ may differ in architecture and capacity; and
(3) \emph{training heterogeneity} -- clients enter the collaborative phase with varying training histories.
Rather than converging to a shared consensus model, each client pursues a \emph{personalized} objective, i.e., it seeks the parameters $\theta_i^\ast$ that
minimize its expected local task loss over its own data distribution:
\begin{equation} \label{eq:local-objective}
    \theta_i^\ast = \operatorname*{arg\,min}_{\theta_i}\;
    \mathbf{E}_{(x, y) \sim D_i}\!\left[\, \mathcal{L}_i^t\!\left(f_{\theta_i}(x), y\right) \right],
\end{equation}
where $f_{\theta_i}$ denotes a forward pass through $\theta_i$ and $\mathcal{L}_i^t$ the task loss.
Because clients hold heterogeneous architectures, it is implied that knowledge cannot be shared by averaging parameters as in weight-aggregation schemes.
We cannot assume that distillation between any two clients would be mutually beneficial, therefore we do not assume knowledge transfers should always be mutual or bi-directional; we assume that they can be asymmetric.

\begin{table}[t]
  \centering
  \caption{Summary of notation.}
  \label{tab:notation}
  \small
  \begin{tabular}{c l}
    \hline
    \textbf{Symbol} & \textbf{Description} \\
    \hline
    $n$              & Number of clients in the network \\
    $M_i$            & Neighbors of client $i$ (candidate peers) \\
    $m_i$              & Number of peers contacted per round, $|\mathcal{C}_i|$ \\
    $\mathcal{C}_i$  & Selected peers of client $i$, $\mathcal{C}_i \subseteq M_i$ \\
    \hline
    $D_i$            & Private local dataset of client $i$ \\
    $\theta_i$       & Native model of client $i$ \\
    $f_{\theta_i}$   & Forward pass of input $x$ through model $\theta_i$ \\
    $y_i$            & Soft outputs of $f_{\theta_i}(...)$ \\
    \hline
    $\mathcal{L}_{i,k}$   & Total (unified) loss for client $i$ w.r.t.\ peer $k$ \\
    $\mathcal{L}^t_{i}$   & Ground-truth task loss of model $\theta_i$ \\
    $\mathcal{L}^d_{i,k}$ & Distillation loss from $\theta_k$ into $\theta_i$ \\
    $||\nabla||_i$        & Gradient (or weight update) norm at client $i$ \\
    $\alpha_t, \alpha_d$  & Task- and distillation-loss signal strengths \\
    \hline
    $\gamma_{i,k}$   & Utility signal (coefficient/score) (Eq.~\ref{eq:akd-coefficient}) \\
    $\gamma_0$       & Default neutral utility signal, $\gamma_0 = 0.5$ \\
    $\Gamma_i(\gamma)$     & $||\nabla||_i$-based modulation of $\gamma$ at client $i$ (Eq.~\ref{eq:mod2}) \\
    $\mathcal{S}_i$        & Score record (map) maintained by client $i$ (Eq.~\ref{eq:score-record-domain}) \\
    $\mathcal{S}_i(k)$     & Recalled score for target peer $k$ by client $i$ (Eq.~\ref{eq:score-recall}) \\
    $P_i$                  & Peer-selection probability distribution \\
    $\phi$                 & Monotonic score-scaling function \\
    \hline
    $\epsilon$             & Local training epochs per round \\
    $\tau_{\mathrm{pre}}$  & Independent local pre-training rounds \\
    $\tau$                 & Collaboration rounds \\
    \hline
  \end{tabular}
\end{table}

\subsection{Decentralized Adaptive Knowledge Distillation} \label{sec:akd}
To enable collaborative training between heterogeneous model architectures, we use \emph{knowledge distillation} as the mechanism to transfer knowledge from a received model into the native model.
In standard distillation-based training, the loss of some \textit{student} model $\theta_i$ given some \textit{teacher} model $\theta_k$ is typically defined as the weighted sum of the ground truth-informed task loss $\mathcal{L}^t_{i}$ and the teacher-based distillation loss  $\mathcal{L}^d_{i, k}$:
\begin{equation}\label{eq:og-kd-loss}
    \mathcal{L}_{i, k} = \alpha_t\mathcal{L}^t_{i} + \alpha_d\mathcal{L}^d_{i, k}
\end{equation}
where $\alpha_t$ and $\alpha_d$ represent the task loss and distillation loss strengths, respectively. The distillation loss $\mathcal{L}^d_{i, k}$ is derived from the Kullback-Leibler (KL) divergence between soft output probabilities of the student and teacher models given a common input $x$ resulting from a forward pass $f$, $y_i = f_{\theta_i}(x)$ and $y_k = f_{\theta_k}(x)$, respectively:
\begin{equation}\label{eq:og-kd-dist-loss}
    \mathcal{L}^d_{i, k} = \text{KL}(y_i, y_k)
\end{equation}
In this manner, the student model is penalized for divergence from the teacher model, and the resulting distillation incentivizes the student model behavior to conform to the teacher model behavior.
Importantly, this naive implementation of this distillation loss assumes that the teacher model has achieved sufficient task performance and is advantageous (i.e., has useful knowledge to offer) to the student model.

\paragraph{Adaptive Knowledge Distillation}
However, for some arbitrary model $\theta_i$ considering distillation from another model from $\theta_k$, it cannot be assumed that $\theta_k$ is always an adequate teacher, and engaging in knowledge distillation with an inadequate teacher model can hurt the overall training of $\theta_i$.
To address this possibility in the case of peer-to-peer learning on networks with heterogeneous data distribution, we must modulate the distillation loss $\mathcal{L}^d_{i, k}$ to be higher, the more advantageous the received model $\theta_k$ is over the native model $\theta_i$ given the local dataset $D_i$.
Thus, to enact this desired behavior
-- i.e., $\mathcal{L}^t_k < \mathcal{L}^t_i$ -- while ensuring that the actuating modulation remains bounded, we use the utility signal:
\begin{equation} \label{eq:akd-coefficient}
    \gamma_{i, k} = \frac{\mathcal{L}^t_{i}}{\mathcal{L}^t_{i} + \mathcal{L}^t_{k}} \in [0, 1]
\end{equation}
As the received model significantly performs better and the native model performs worse ($\mathcal{L}^t_{k}\to 0$ and $\mathcal{L}^t_{i}\to\infty$), the utility signal increases $\gamma_{i, k}\to 1$, thus promoting the distillation loss.
When the received model performs worse and the native model performs better ($\mathcal{L}^t_{k}\to\infty$ and $\mathcal{L}^t_{i}\to 0$), the coefficient decreases $\gamma_{i, k}\to 0$, thus suppressing the distillation loss.
In the neutral case that both models perform about the same, $\gamma_{i, k}\to 0.5$.

\paragraph{Modulating the Modulation}
While we can use this utility signal directly as a coefficient to scale distillation loss based on the inferred utility of a candidate peer, the subsequent selectivity would hurt the generalized knowledge building that takes place in earlier collaborative training rounds~\cite{fl-general}.
Therefore, we further modulate the utility signal itself by scaling it based on gradient (or weight update) norms, such that the relationship between the client $i$'s recent gradient norm $||\nabla||_i^{recent}$ and the maximum historical gradient norm $||\nabla||_i^{max}$ interpolates the final coefficient between the utility signal coefficient and a full signal $\gamma_{\mathrm{full}} = 1$:
\begin{equation}\label{eq:mod2}
    \Gamma_i(\gamma) = \left(1 - \frac{||\nabla||_i^{recent}}{||\nabla||_i^{max}}\right) \gamma\, + \, \frac{||\nabla||_i^{recent}}{||\nabla||_i^{max}}\, \in [0, 1]
\end{equation}
With this, when a client is in earlier stages of training and gradient updates are larger ($\frac{||\nabla||_i^{recent}}{||\nabla||_i^{max}} \rightarrow 1$), it will not differentiate its modulation among contributed distillation losses, $\Gamma(\gamma) \rightarrow 1$.
When a client is in later stages of training and gradient updates are smaller ($\frac{||\nabla||_i^{recent}}{||\nabla||_i^{max}} \rightarrow 0$), it will retain full differentiation through the per-peer-$k$ utility signal, $\Gamma_i(\gamma_{i, k}) \rightarrow \gamma_{i, k}$.
Finally, we define our unified loss function with adaptive knowledge distillation as:
\begin{equation} \label{eq:akd-unified-loss}
    \mathcal{L}_{i, k} = \alpha_t\mathcal{L}^t_{i} + \alpha_d\,\Gamma_i\left(\gamma_{i,k}\right)\,\text{KL}(y_i, y_k)
\end{equation}

\paragraph{Decentralized Multi-Distillation}
We consider that a client $i$ may receive models from $m_i \geq 1$ peers $\mathcal{C}_i \subseteq M_i$ at once or accumulates them over time before engaging in distillation.
Then, to facilitate distillation into student model $\theta_i$ from multiple teacher models $\theta_k\ \forall\ k \in \mathcal{C}_i$ simultaneously, we average all received models' contributed distillation losses into an aggregate distillation loss:
\begin{equation} \label{eq:dkd-loss}
    \mathcal{L}_{i, k} = \alpha_t\mathcal{L}^t_{i} + \frac{\alpha_d}{m_i}\sum^{\mathcal{C}_i}_k\mathcal{L}^d_{i, k}
\end{equation}
Recall that each contributed distillation is weighed by the modulated coefficient $\Gamma_i(\gamma_{i, k})$ (Eq.~\ref{eq:mod2}), yielding:
\begin{equation} \label{eq:complete-dakd-loss}
    \mathcal{L}_{i, k} = \alpha_t\mathcal{L}^t_{i} + \frac{\alpha_d}{m_i}\sum^{\mathcal{C}_i}_k\Gamma_i\left(\gamma_{i,k}\right)\text{KL}(y_i, y_k)
\end{equation}

\subsection{Utility-based Peer Selection} \label{sec:ps}

In our case of asymmetric decentralized knowledge distillation, the optimal peer selection process at client $i$ should result in model update requests from connected clients $k \in M$ whose models $\theta_{k \in M}$ can best contribute via distillation to the local model $\theta_i$.
To that end, we posit that we can derive this contribution potential based on the relative task loss advantage of any model $\theta_k$ over $\theta_i$.
Our \emph{adaptive distillation} already yields the utility signal and adaptive coefficient using those task losses, $\gamma_{i, k} = \frac{\mathcal{L}^t_{i}}{\mathcal{L}^t_{i} + \mathcal{L}^t_{i}}$ as a by-product.
Thus, for the same reason we apply this task loss advantage to adaptive distillation, we can utilize $\gamma$ as a score value bounded $[0, 1]$ such that $\gamma_{i,k} = 0$ implies no contribution potential, $\gamma_{i,k} = 1$ implies strong contribution potential, and $\gamma_{i,k} = 0.5$ implies neutral contribution potential.

\paragraph{Maintaining Scores}
To leverage these scores for peer selection, each client $i$ maintains its own score record $\mathcal{S}_i$, a map that is updated with each model transfer from another client, such that $\mathcal{S}_{i,k}$ contains the score $\gamma_{i,k}$ from its most recent encounter with client $k$ for only those clients that client $i$ has encountered.
Then, letting $\mathcal{K}$ be the key space for any $\mathcal{S}$ and $\mathrm{dom}(\mathcal{S}_{i})$ be the set of all recorded clients:
\begin{equation} \label{eq:score-record-domain}
    \mathcal{S}: \mathcal{K} \rightharpoonup \gamma \in [0,1] \text{ and } \mathrm{dom}(\mathcal{S}_{i}) \subseteq \mathcal{K}_i \subseteq M_i
\end{equation}
When client $i$ recalls its score for some client $k$, $\mathcal{S}_i(k)$, the most recent score is obtained if it has encountered $k$, else the score is assumed to be the default neutral score $\gamma_0 = 0.5$, such that:
\begin{equation} \label{eq:score-recall}
    \mathcal{S}_i(k) =
    \begin{cases}
        \mathcal{S}_{i,k} \mapsto \gamma_{i,k} & \text{if } k \in \mathrm{dom}(\mathcal{S}_{i}) \\
        \gamma_0 = 0.5 & \text{if } k \notin \mathrm{dom}(\mathcal{S}_{i})
    \end{cases}
\end{equation}

\paragraph{Optimal Peer Selection}
For client $i$ to select $m$ optimal peers $\mathcal{C}_i \subseteq M_i$, it leverages the recalled scores $\mathcal{S}_i(k)\ \forall\ k \in M_i$ to model selection as a multi-armed bandit problem.
We can use score-proportionate selection -- weighted sampling without replacement (w.o.r.), i.e., clients with higher scores have a higher chance of being selected -- with an optional monotonic increasing scaling function $\phi$ being applied to the scores to manipulate the probability distribution:
\begin{equation}
    \mathcal{C}_i \sim \operatorname*{multinomial}^{m \text{, w.o.r.}}_{k \in M_i}(P_i) \text{, }
    P_i(k) = \frac{\phi\bigl(\mathcal{S}_i(k)\bigr)}{\sum_{j \in M_i} \phi\bigl(\mathcal{S}_i(j)\bigr)}
\end{equation}
Or, we can lean towards exploitation over exploration use greedy selection to select clients with the top $m$ scores:
\begin{equation} \label{eq:greedy-ps}
    \mathcal{C}_i = \operatorname*{arg\,max}^m_{k \in M_i} \; \mathcal{S}_i(k)
\end{equation}
Other multinomial bandit solvers may also be used as well, but a thorough exploration of those solutions is outside the scope of this paper.

\begin{algorithm}[t]
\caption{SPADE-KD, at client $i$}
\label{alg:spade-kd}
\textbf{Input}: local dataset $D_i$; native model $\theta_i$; candidate peers $M_i \in \mathbf{N}$ \\
\textbf{Parameters}: peers per round $m_i$; local epochs $\epsilon$; pre-training rounds $\tau_{\mathrm{pre}}$; collaboration rounds $\tau$; task and distillation loss strengths $\alpha_t, \alpha_d$; default utility signal $\gamma_0$; learning rate $\eta$ \\
\textbf{Output}: personalized model $\theta_i$
\begin{algorithmic}[1]
\STATE Initialize score record $\mathcal{S}_i \gets \emptyset$, where\\
$\mathcal{S}_i(k) = \gamma_0$ if $k$ is uncontacted, else $\mathcal{S}_{i,k}$
\STATE \textit{// Pre-training phase}
\FOR{round $= 1$ \TO $\tau_{\mathrm{pre}}$}
    \STATE $\theta_i \gets \epsilon$ epochs of training on $D_i$ minimizing $\mathcal{L}^t_{i}$
\ENDFOR 
\STATE \textit{// Collaboration phase}
\FOR{round $= 1$ \TO $\tau$}
    \STATE $\theta_i \gets \epsilon$ epochs of training on $D_i$ minimizing $\mathcal{L}^t_{i}$
    \STATE \textit{// Utility-based peer selection}
    \STATE $\mathcal{C}_i \gets \operatorname*{arg\,max}^{\,m_i}_{k \in M_i}\ \mathcal{S}_i(k)$ \COMMENT{or sample $\propto \phi(\mathcal{S}_i(k))$}
    \STATE Request and receive $\{\theta_k : k \in \mathcal{C}_i\}$
    \STATE \textit{// Decentralized adaptive knowledge distillation}
    \FOR{$\epsilon$ epochs}
        \FORALL{inputs $x \subset D_i$}
            \STATE $y_i \gets f_{\theta_i}(x)$;\, Evaluate $\mathcal{L}^t_{i}$ on $D_i$ 
            \STATE \textit{// Compute $\gamma$ signal, store recent as score}
            \FORALL{$k \in \mathcal{C}_i$}
                \STATE $y_k \gets f_{\theta_k}(x)$;\, Evaluate $\mathcal{L}^t_{k}$ on $D_i$
                \STATE $\gamma_{i,k} \gets \dfrac{\mathcal{L}^t_{i}}{\mathcal{L}^t_{i} + \mathcal{L}^t_{k}}$
                \STATE $\mathcal{S}_{i,k} \gets \gamma_{i,k}$
            \ENDFOR
            \STATE $\displaystyle \mathcal{L}_i \gets \alpha_t \mathcal{L}^t_{i} + \frac{\alpha_d}{m_i}\sum_{k \in \mathcal{C}_i} \Gamma_i(\gamma_{i,k})\,\mathrm{KL}(y_i, y_k)$
            \STATE $\theta_i \gets \theta_i - \eta\, \nabla_{\theta_i} \mathcal{L}_i$
        \ENDFOR
    \ENDFOR
\ENDFOR
\STATE \textbf{return} $\theta_i$
\end{algorithmic}
\end{algorithm}

%% file: sections/4_experiment.tex
\section{Experiments}

In our experiments, we implement and compare SPADE-KD and its ablations alongside the naive baseline.
We measure performance in terms of loss in different classification task scenarios under fully heterogeneous settings.

\subsection{Scenario Configurations}

We use a custom-built discrete event simulator based on Python 3 to simulate synchronous rounds of distributed learning with models being implemented with PyTorch.
Simulations were run on an Apple Mac Studio with an Apple M4 Max chip with 16 cores and 128 GB of memory.
Each decentralized paradigm is run considering a peer-to-peer small-world network~\cite{watts-strogatz} of $n$ clients, where each client can collaborate with $m$ other partner clients each round.
Clients complete $\tau_{\mathrm{pre}}$ rounds, with each round consisting of $\epsilon$ epochs, of purely local pre-training before the collaborative phase begins.
Collaborative learning proceeds for $\tau$ rounds, and in each round an agent performs $\epsilon$ epochs of local training on its private data before exchanging knowledge with its partners.

\paragraph{Data Heterogeneity}
We consider classification task scenarios built around 3 distributed datasets -- \textit{MNIST}~\cite{mnist}, \textit{FEMNIST}~\cite{femnist}, and \textit{SpeechCommands}~\cite{speechcommandsv2} -- with the latter two being provided by Flower~\cite{flower-datasets}.
Table~\ref{tab:exp-configs} shows the configuration parameters for each scenario.
In the \textit{MNIST} scenario ($\sim$600 samples/client), we introduce artificial heterogeneity by \textit{label-swapping}, i.e., swapping unique sets of labels within each data distribution~\cite{sim-effects}.
We use this scenario to represent a more extreme case of heterogeneity expressed through label incongruence, whereas the \textit{FEMNIST} scenario presents the same task with natural heterogeneity due to its client partitions being based on the original writers of the digits ($\sim$53 samples/client).
The \textit{SpeechCommands} dataset is also naturally heterogeneous with its client partitions being based on the original speakers of the words, but resembles more massively distributed real-world scenarios in which each client holds very little data ($\sim$25 samples/client). 

\begin{table}[t]
  \centering
  \caption{Experimental configuration for each scenario.
    }
  \label{tab:exp-configs}
  \begin{tabular}{l|ccccc}
    Scenario & $n$ & $m$ & $\epsilon$ & $\tau_{\mathrm{pre}}$ & $\tau$\\ \hline
    \textit{MNIST} & 100 & 10 & 3 & $\leq$10 & 40 \\
    \textit{FEMNIST} & 80 & 10 & 20 & $\leq$40 & 60 \\
    \textit{SpeechCommands} & 80 & 5 & 2 & $\leq$15 & 60 \\
  \end{tabular}
\end{table}

\paragraph{Model \& Training Heterogeneity}
We model deep learning model heterogeneity by implementing size-based tiers within the model family for each task.
For the \textit{MNIST} and \textit{FEMNIST} handwritten digit recognition task scenarios, we employ a family of four CNN classifiers.
The largest is the standard LeNet-5 ($\sim$61K parameters), while the three more compact variants ($\sim$29K, $\sim$12.6K, and $\sim$3.4K parameters) share a common single-convolution template -- a $3\times3$ convolutional layer with $4\times4$ max-pooling followed by a single fully-connected hidden layer -- with the channel width and hidden dimension scaled down at each tier.
For the \textit{SpeechCommand} keyword spotting task scenario we use a shared keyword transformer architecture~\cite{kwt} instantiated at three capacity tiers ($\sim$76K, $\sim$32K, and $\sim$7K parameters).
We select these small models to represent models that can be trained on commonly available edge devices and to represent the effects of models with varying capacity in collaboration with each other.
At the start of each experiment, model architectures are assigned round-robin to each client and clients pre-train their model on local data at different levels, randomly set.

\input{plots/build-comparison-table}

\subsubsection{Method Implementation}

To our knowledge, no fully decentralized approach provides a model-agnostic approach to this high-heterogeneity problem space where clients distill from multiple peer collaborators simultaneously.
Thus, to observe the relative advantage of our approach -- wholly and with each of its component mechanisms -- against a fully decentralized baseline, we implement the following\footnote{Comparison against the related distributed distillation method(s) is included in the supplemental appendix and is omitted here due to the infeasibility of direct comparison between different collaboration structures.}:
\begin{itemize}
    \item \textbf{Decentralized KD} (D-KD)\\naive baseline implementation of decentralized KD (Eq.~\ref{eq:dkd-loss}) with the standard distillation loss (Eq.~\ref{eq:og-kd-dist-loss})
    \item \textbf{Decentralized KD + Adaptive Distillation} (D-AKD)\\decentralized KD implemented with the adaptive distillation loss signal (Eq.~\ref{eq:complete-dakd-loss})
    \item \textbf{Decentralized KD + Peer Selection} (D-KD-PS)\\decentralized KD implemented with the greedy utility-driven peer selection (Eq.~\ref{eq:greedy-ps})
    \item \textbf{SPADE-KD} ($\spadesuit$-KD)\\our complete decentralized KD approach (Alg.~\ref{alg:spade-kd}) implemented with both adaptive knowledge distillation and greedy utility-based peer selection
\end{itemize}
In our experiments, we want to observe the collaboration advantage, i.e., how beneficial distillation is across different approaches.
Therefore, since clients already perform local training at each round, we set $\alpha_t = 0$ and $\alpha_d = 1$ to completely suppress the supervision task loss while retaining the full distillation loss during the distillation step.
We opt to use greedy peer selection (Eq.~\ref{eq:greedy-ps}) to best show the impact of our utility-based peer selection.

\subsection{Results \& Discussion}

\input{plots/build-graphs}

We monitor loss as our primary performance metric\footnote{Accuracies are included in the supplemental appendix.}.
Results are shown for the \textit{FEMNIST} scenario in Fig.~\ref{res:femnist}, the label-swapped \textit{MNIST} scenario in Fig.~\ref{res:mnist}, and the \textit{SpeechCommands} scenario in Fig.~\ref{res:speech}.
Final performance losses and relative performance advantages are shown in Table~\ref{tab:summary}.
These results reflect the performance metrics of all clients averaged across 5 unique seeds for the \textit{MNIST} and \textit{FEMNIST} scenarios and 10 unique seeds for the \textit{SpeechCommands} scenario to account for noisiness incurred from scare data availability.
Our experiments demonstrate that overall, while both its adaptive distillation and utility-based peer selection components each provide tangible performance advantages, the combination of both in SPADE-KD yields better performance than each component in isolation in virtually all cases.

\paragraph{Component Effectiveness}
The impacts of each component mechanism do vary.
Earlier in the collaboration phase (as seen in the \textit{FEMNIST} or \textit{MNIST} scenarios) or when dealing with scarce local data availability (as seen in the \textit{SpeechCommands} scenario), adaptive distillation without utility-based peer selection tends to be the more advantageous component, achieving higher initial performance gains than peer selection without adaptive distillation.
However, as collaborative training continues and native models become more stationary, utility-based peer selection without adaptive distillation seems to provide a greater advantage than adaptive distillation without utility-based peer selection.
We can even observe that in the \textit{FEMNIST} and \textit{MNIST} scenarios, utility-based peer selection in isolation appears to match or slightly exceed the accuracy of SPADE-KD; however, even in these cases, SPADE-KD still achieves lower overall loss.

\paragraph{Managing Quality Disparity}
We observe that our approach yields the lowest loss for both higher-quality models (via $P_{10}$ losses) and lower-quality models (via $P_{90}$ losses) in virtually all cases.
This implies that our method is effective in suppressing the distillation influence of lower-quality models and preserving the training progress of higher-quality models while still enabling the unidirectional flow of knowledge to those lower-quality models.
We observe a specific exception in the massively distributed \textit{SpeechCommands} scenario -- our approach does not serve lower-quality client models as well (Fig.~\ref{res:speech-loss-p90}); we can attribute this to the stochastic noise expected in the weakest models training over the fewest local samples, but this leaves room for future exploration.

\paragraph{Diminishing Scores}
We observe that clients' score values for contacted peers decrease over time.
This makes sense -- while contribution potential is high in earlier rounds of training, as a client has more rounds to train on its local data, the potential for another client's model trained on that client's data diminishes.
This implicit decaying of score also acts as a natural defense against stale score values, such that a peer with a stale high score may be selected for knowledge transfer but that interaction would update that stale score.
Even then, optimal peer selection still selects the most useful peers to request models from, and as scores diminish, clients are encouraged to contact unexplored peers due to the default neutral contribution potential $\gamma_0 = 0.5$.



%% file: plots/build-comparison-table.tex
%
%
%

\begin{filecontents*}{plots/data/femnist_comparison.csv}
scenario,method,column,round,mean_loss,mean_impr_pct,p10_loss,p10_impr_pct,p90_loss,p90_impr_pct
femnist,D-KD,DecentralizedKD,60,0.619609,0.0,0.346189,0.0,0.969468,0.0
femnist,D-KD-PS,DecentralizedKD_PS,60,0.565978,8.66,0.313705,9.38,0.89031,8.17
femnist,P2P-AKD,P2P_AKD,60,0.529401,14.56,0.252592,27.04,0.905281,6.62
femnist,SPADE-KD,SpadeKD,60,0.54755,11.63,0.276332,20.18,0.894166,7.77
femnist,D-AKD,DecentralizedA2KD,60,0.586472,5.35,0.299518,13.48,0.95276,1.72
femnist,D-AKD-old,DecentralizedAKD,60,0.548599,11.46,0.253959,26.64,0.938683,3.18
\end{filecontents*}
\DTLloaddb{cmpFem}{plots/data/femnist_comparison.csv}
\begin{filecontents*}{plots/data/mnist_comparison.csv}
scenario,method,column,round,mean_loss,mean_impr_pct,p10_loss,p10_impr_pct,p90_loss,p90_impr_pct
mnist,D-KD-PS,DecentralizedKD_PS,40,0.344046,20.36,0.18475,17.36,0.526901,23.78
mnist,P2P-AKD,P2P_AKD,40,0.331644,23.23,0.160995,27.98,0.525015,24.05
mnist,D-AKD,DecentralizedA2KD,40,0.398958,7.65,0.203625,8.91,0.655142,5.22
mnist,D-KD,DecentralizedKD,40,0.432007,0.0,0.223547,0.0,0.69125,0.0
mnist,SPADE-KD,SpadeKD,40,0.33256,23.02,0.167571,25.04,0.523576,24.26
mnist,D-AKD-old,DecentralizedAKD,40,0.370243,14.3,0.172506,22.83,0.621692,10.06
\end{filecontents*}
\DTLloaddb{cmpMni}{plots/data/mnist_comparison.csv}
\begin{filecontents*}{plots/data/speech_comparison.csv}
scenario,method,column,round,mean_loss,mean_impr_pct,p10_loss,p10_impr_pct,p90_loss,p90_impr_pct
speech,D-KD-PS,DecentralizedKD_PS,60,1.085947,0.87,0.383181,2.61,1.820869,-0.94
speech,D-AKD,DecentralizedA2KD,60,1.075052,1.86,0.352128,10.5,1.826152,-1.24
speech,D-KD,DecentralizedKD,60,1.09545,0.0,0.393436,0.0,1.803849,0.0
speech,SPADE-KD,SpadeKD,60,1.072557,2.09,0.348643,11.39,1.830869,-1.5
\end{filecontents*}
\DTLloaddb{cmpSpe}{plots/data/speech_comparison.csv}

\foreach \db/\pfx in {cmpFem/Fem, cmpMni/Mni, cmpSpe/Spe}{%
  \DTLgetvalueforkey{\tmp}{round}{\db}{method}{D-KD}\expandafter\xdef\csname Round\pfx\endcsname{\tmp}%
  \foreach \m/\mid in {D-KD/DKD, D-KD-PS/DKDPS, D-AKD/DAKD, SPADE-KD/SPADE}{%
    \foreach \st in {mean, p10, p90}{%
      \DTLgetvalueforkey{\tmp}{\st_loss}{\db}{method}{\m}\expandafter\xdef\csname L\pfx\mid\st\endcsname{\tmp}%
      \DTLgetvalueforkey{\tmp}{\st_impr_pct}{\db}{method}{\m}\expandafter\xdef\csname P\pfx\mid\st\endcsname{\tmp}%
    }%
  }%
}%

\newcommand{\cmploss}[3]{
  \begingroup\edef\rawL{\csname L#1#2#3\endcsname}%
  \pgfmathprintnumber[fixed,fixed zerofill,precision=3]{\rawL}\endgroup}
\newcommand{\cmppct}[3]{
  \begingroup
  \edef\rawP{\csname P#1#2#3\endcsname}\def\midarg{#2}\def\dkdc{DKD}%
  \ifx\midarg\dkdc$-$\else\textit{\pgfmathprintnumber[fixed,precision=1,print sign,fixed zerofill]{\rawP}}\fi
  \endgroup}
\newcommand{\cmprow}[2]{
  #1
  & \cmploss{Fem}{#2}{mean} & \cmppct{Fem}{#2}{mean} & \cmploss{Fem}{#2}{p10} & \cmppct{Fem}{#2}{p10} & \cmploss{Fem}{#2}{p90} & \cmppct{Fem}{#2}{p90}
  & \cmploss{Mni}{#2}{mean} & \cmppct{Mni}{#2}{mean} & \cmploss{Mni}{#2}{p10} & \cmppct{Mni}{#2}{p10} & \cmploss{Mni}{#2}{p90} & \cmppct{Mni}{#2}{p90}
  & \cmploss{Spe}{#2}{mean} & \cmppct{Spe}{#2}{mean} & \cmploss{Spe}{#2}{p10} & \cmppct{Spe}{#2}{p10} & \cmploss{Spe}{#2}{p90} & \cmppct{Spe}{#2}{p90} \\}

\begin{table*}[t]
  \centering
  \caption{Final-round client loss ($\mathcal{L}$) and percentage improvement ($\Delta\%$) of the utility-based peer selection ablation (D-KD-PS), the adaptive knowledge distillation ablation (D-AKD), and the complete SPADE-KD ($\spadesuit$-KD) method versus the naive decentralized KD (D-KD) baseline, at the mean, $P_{10}$, and $P_{90}$ across clients.}
  \label{tab:summary}
  \footnotesize
  \setlength{\tabcolsep}{2pt}
  \begin{tabular}{|l|rr|rr|rr|rr|rr|rr|rr|rr|rr|}
    \hline
    & \multicolumn{6}{c|}{\textit{FEMNIST}}
    & \multicolumn{6}{c|}{\textit{MNIST}}
    & \multicolumn{6}{c|}{\textit{SpeechCommands}}\\
    & \multicolumn{2}{c|}{Mean} & \multicolumn{2}{c|}{$P_{10}$} & \multicolumn{2}{c|}{$P_{90}$}
    & \multicolumn{2}{c|}{Mean} & \multicolumn{2}{c|}{$P_{10}$} & \multicolumn{2}{c|}{$P_{90}$}
    & \multicolumn{2}{c|}{Mean} & \multicolumn{2}{c|}{$P_{10}$} & \multicolumn{2}{c|}{$P_{90}$}\\
    Method & \multicolumn{1}{c}{$\mathcal{L}$} & \multicolumn{1}{c|}{$\Delta\%$} & \multicolumn{1}{c}{$\mathcal{L}$} & \multicolumn{1}{c|}{$\Delta\%$} & \multicolumn{1}{c}{$\mathcal{L}$} & \multicolumn{1}{c|}{$\Delta\%$} & \multicolumn{1}{c}{$\mathcal{L}$} & \multicolumn{1}{c|}{$\Delta\%$} & \multicolumn{1}{c}{$\mathcal{L}$} & \multicolumn{1}{c|}{$\Delta\%$} & \multicolumn{1}{c}{$\mathcal{L}$} & \multicolumn{1}{c|}{$\Delta\%$} & \multicolumn{1}{c}{$\mathcal{L}$} & \multicolumn{1}{c|}{$\Delta\%$} & \multicolumn{1}{c}{$\mathcal{L}$} & \multicolumn{1}{c|}{$\Delta\%$} & \multicolumn{1}{c}{$\mathcal{L}$} & \multicolumn{1}{c|}{$\Delta\%$}\\

    \hline
    \cmprow{D-KD}{DKD}
    \cmprow{\textit{D-KD-PS}}{DKDPS}
    \cmprow{\textit{D-AKD}}{DAKD}
    \cmprow{\textbf{$\spadesuit$-KD}}{SPADE}
    \hline
  \end{tabular}
\end{table*}

%% file: plots/build-graphs.tex
\newcommand{\progRounds}{}
\newcommand{\progMetric}{}

\newcommand{\subfigWidth}{0.32\textwidth}

\newif\iffedkd
\fedkdtrue

\newcommand{\lossPanel}[3]{%
\begin{subfigure}[b]{\subfigWidth}
    \centering
    \begin{tikzpicture}
        \begin{axis}[
            font=\scriptsize,
            ylabel={\textsc{\progMetric}},
            xlabel={\textsc{rounds}},
            xlabel near ticks,
            xmin=1, xmax=\progRounds,
            width=\linewidth,
            height=\linewidth,
            grid=major,
        ]
            \iffedkd
            \addplot[color=gray!80!black, mark=none]        table [x expr=\coordindex + 1, y=FedKD,              col sep=comma, header=true] {#1};
            \fi
            \addplot[color=blue, mark=none]                 table [x expr=\coordindex + 1, y=DecentralizedKD,    col sep=comma, header=true] {#1};
            \addplot[color=teal, mark=none]                 table [x expr=\coordindex + 1, y=DecentralizedKD_PS, col sep=comma, header=true] {#1};
            \addplot[color=violet, mark=none]               table [x expr=\coordindex + 1, y=DecentralizedA2KD,  col sep=comma, header=true] {#1};
            \addplot[color=orange, mark=none, line width=1.1pt] table [x expr=\coordindex + 1, y=SpadeKD,         col sep=comma, header=true] {#1};
        \end{axis}
    \end{tikzpicture}
    \caption{#2}
    \label{#3}
\end{subfigure}%
}

\newcommand{\legendItem}[2]{%
    \tikz[baseline=-0.5ex]{\draw[#1] (0,0) -- (0.45,0);}~#2}
\newcommand{\methodsLegend}{{\scriptsize
    \iffedkd\legendItem{gray!80!black, line width=1pt}{FedKD}\quad\fi
    \legendItem{blue,          line width=1pt}{D-KD}\quad
    \legendItem{teal,          line width=1pt}{D-KD-PS}\quad
    \legendItem{violet,        line width=1pt}{D-AKD}\quad
    \legendItem{orange,        line width=1.1pt}{SPADE-KD}%
}}

\newcommand{\buildGraph}[9]
{%
\renewcommand{\progRounds}{#1}
\renewcommand{\progMetric}{#2}
\lossPanel{#3_mean.csv}{#4}{#5}\hfill
\lossPanel{#3_p10.csv}{#6}{#7}\hfill
\lossPanel{#3_p90.csv}{#8}{#9}%
\\[0.4em]
\methodsLegend
}

\fedkdfalse
\begin{figure*}
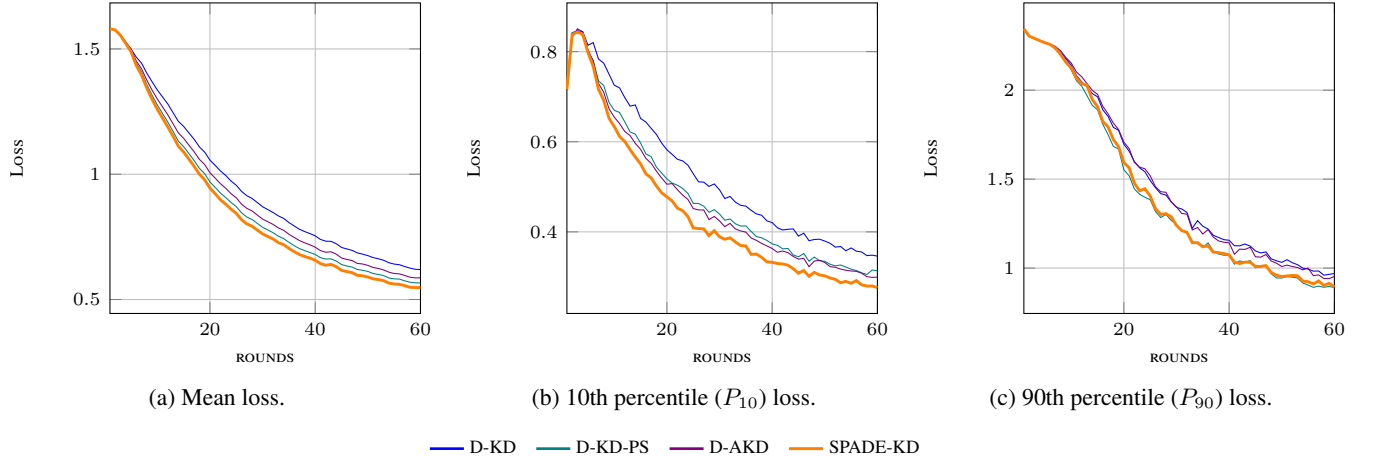

    \centering
    \buildGraph{60}{Loss}{plots/data/femnist_loss}
        {Mean loss.}{res:femnist-loss-mean}
        {10th percentile ($P_{10}$) loss.}{res:femnist-loss-p10}
        {90th percentile ($P_{90}$) loss.}{res:femnist-loss-p90}
    \caption{Client loss on the \textit{FEMNIST} scenario (naturally heterogeneous image data): mean, $P_{10}$, and $P_{90}$ across clients.}
    \label{res:femnist}
\end{figure*}

\begin{figure*}
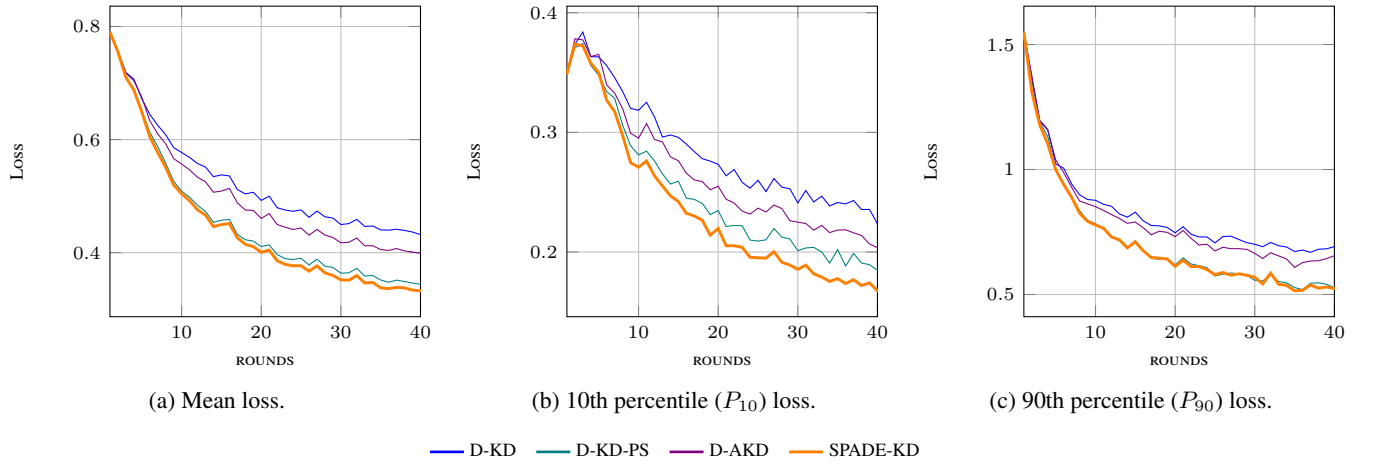

    \centering
    \buildGraph{40}{Loss}{plots/data/mnist_loss}
        {Mean loss.}{res:mnist-loss-mean}
        {10th percentile ($P_{10}$) loss.}{res:mnist-loss-p10}
        {90th percentile ($P_{90}$) loss.}{res:mnist-loss-p90}
    \caption{Client loss on the \textit{MNIST} scenario (artificially heterogeneous image data): mean, $P_{10}$, and $P_{90}$ across clients.}
    \label{res:mnist}
\end{figure*}

\begin{figure*}
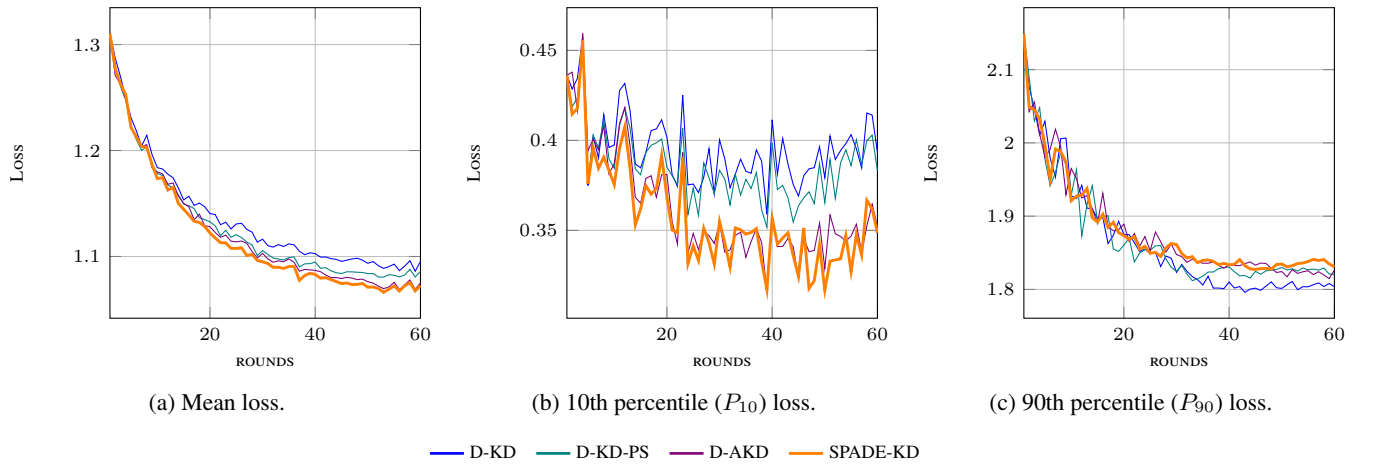

    \centering
    \buildGraph{60}{Loss}{plots/data/speech_loss}
        {Mean loss.}{res:speech-loss-mean}
        {10th percentile ($P_{10}$) loss.}{res:speech-loss-p10}
        {90th percentile ($P_{90}$) loss.}{res:speech-loss-p90}
    \caption{Client loss on the \textit{SpeechCommands} scenario (naturally heterogeneous audio data): mean, $P_{10}$, and $P_{90}$ across clients.}
    \label{res:speech}
\end{figure*}

%% file: sections/5_conclusion.tex
\section{Conclusion}

We presented SPADE-KD, a fully decentralized, serverless framework for collaborative learning among clients that are heterogeneous in their data distributions, model architectures, and model quality.
Its core idea is that a single, locally computed utility signal --- the relative task-loss advantage of a received model over a client's own model on its local data --- can act as an adaptive coefficient for how much a client distills from a received model and as a bandit score for which candidate peers it queries next.
Across three classification scenarios spanning artificial and natural heterogeneity, adaptive distillation and utility-based peer selection each yielded tangible advantages, and their combination in SPADE-KD generally achieved the lowest loss --- surpassing a the naive decentralized distillation baseline and either component in isolation.
Our future work will focus on adapting distillation selectivity to a client's data scarcity or training stage and exploring richer multi-armed bandit formulations for peer selection beyond the policies considered here.

%% file: appendix/6_appendix.tex
\clearpage
\appendix
\renewcommand{\appendixpagename}{Appendix}
\appendixpage
\begin{appendices}

\section{Full Experiment Results}
\input{appendix/build_scores_graph}
\input{appendix/build-graphs}
\input{appendix/build-comparison-table}

We include more comprehensive results of our experiments:
\begin{itemize}
    \item \textbf{Figure~\ref{fig:scores-femnist}}: Scores of candidate peers in SPADE-KD for the \textit{FEMNIST} scenario.
    \item \textbf{Figure~\ref{fig:scores-mnist}}: Scores of candidate peers in SPADE-KD for the \textit{MNIST} scenario.
    \item \textbf{Figure~\ref{fig:scores-speech}}: Scores of candidate peers in SPADE-KD for the \textit{SpeechCommands} scenario.
    \item \textbf{Figure~\ref{res:app-femnist}}: Clients' loss and accuracy for the \textit{FEMNIST} scenario.
    \item \textbf{Figure~\ref{res:app-mnist}}: Clients' loss and accuracy for the label-swapped \textit{MNIST} scenario.
    \item \textbf{Figure~\ref{res:app-speech}}: Clients' loss and accuracy for the \textit{SpeechCommands} scenario.
    \item \textbf{Table~\ref{tab:app-comparison-loss}}: Summary of client performance via final-round loss for all scenarios.
    \item \textbf{Table~\ref{tab:app-comparison-acc}}: Summary of client performance via final-round accuracy for all scenarios.
\end{itemize}

\paragraph{Visualizing Diminishing Scores}

Figs.~\ref{fig:scores-femnist}, \ref{fig:scores-mnist}, and \ref{fig:scores-speech} present average of clients' recorded scores for selected candidate peers as part of the utility-based peer selection in SPADE-KD each scenario to showcase how clients' recorded scores for candidate peers diminishes over time.
Note that at the start of the collaboration phase, however, all clients $i \in \mathbf{N}$ record $\gamma_{i,k} = 0.5$ for all candidate peers $k \in M_i$, since no collaboration has occurred yet between clients and the recalled score is defaulted to $\gamma_0 = 0.5$.
 
\paragraph{Analyzing Accuracy Performance}
We present accuracy performance across the different methods in addition to loss performance in Figs.~\ref{res:app-femnist}, \ref{res:app-mnist}, and \ref{res:app-speech}.
However, we note that the performance advantages in terms of accuracy of our proposed approach SPADE-KD and its ablations over the baselines are harder to discern that in terms of loss.
While it still holds that SPADE-KD achieves the highest accuracies in most cases, it is sometimes matched or surpassed by the utility-based peer selection alone (D-KD-PS); and yet, SPADE-KD still yields the lowest loss.
We infer that while utility-based peer selection alone can yield good prediction accuracy, the confidence of those predictions can be hindered by divergent client drift, whereas adaptive distillation alongside optimal peer selection can reduce divergent client drift and improve loss.
This inference extends beyond the peer selection ablation --- when compared to naive decentralized KD (D-KD) in the \textit{SpeechCommands} scenario, most of the other tested methods appear to achieve worse prediction accuracy, yet achieve lower loss (with the noted exception of the lower-quality models represented by the $P_{90}$ quantile).

\paragraph{Comparing to Federated KD}
The class of fully-decentralized distributed knowledge distillation approaches is still sparse at the time of writing, and the sub-class of \textit{heterogeneity-aware} approaches is even more so, making it difficult to find related works to use as state-of-the-art baselines in our experiments.
Existing semi-decentralized distributed knowledge distillation approaches rely on electing aggregator nodes~\cite{dfml} mutually distill on local data (with the implicit assumption that any heterogeneity in underlying data distributions can be overcome by functional consensus) or centralized matchmakers for pairwise model exchanges~\cite{knexafl}.
In these results, we opt to include FedKD~\cite{fedkd} as related method, since it uses mutual distillation between a publicly aggregated \textit{mentee} model and locally trained and distilled \textit{mentor} model to facilitate adaptive distillation that acknowledges heterogeneity in underlying data distributions.

However, even then, a direct comparison between FedKD and the decentralized distillation approaches we present poses various incompatibilities.
First, this federated approach assumes a different communication structure altogether, where clients interact only with the server; thus, we implement a star topology with the server at the middle instead of the small-world peer-to-peer topology~\cite{watts-strogatz}.
Second, FedKD uses mutual distillation between its mentor and mentee models to transfer knowledge from the rest of the network, whereas our approaches implement distinct local training and collaborative distillation steps.
To attempt to mitigate this, we set $\alpha_t = \alpha_d = 1$ to retain both the full task loss and distillation losses.
Third, FedKD and other federated knowledge distillation approaches~\cite{proxyfl} rely on compact proxy models to carry information between clients through server-orchestrated parameter aggregation, making their performance dependent on the structure and capacity of the proxy model architecture itself.
We consider that in real world scenarios considering a network of heterogeneous model architectures and inferred heterogeneous computational capacities, the proxy model architecture should be trainable by all clients.
Therefore, we select the smallest common model architecture for the intermediary proxy model.
Even then, the existence of these factors, however mitigated, and others make direct comparison challenging between federated KD and our decentralized KD methods.
Future work will compare our methods and ablations against state-of-the-art decentralized heterogeneity-aware distillation methods as they continue to emerge.

\section{Reproducibility \& Code Supplement}

The code supplement has been submitted separately as a compressed \texttt{.zip} file which must be extracted before use.
Instructions for understanding the implementations and reproducing the experiments are located in \texttt{AAAI\_README.md} of the root directory.

\end{appendices}

%% file: appendix/build_scores_graph.tex
%
%
%

\newcommand{\scoreGraph}[2]{
    \begin{tikzpicture}
        \begin{axis}[
            font=\scriptsize,
            xlabel={\textsc{round}},
            ylabel={\textsc{score}},
            xlabel near ticks, ylabel near ticks,
            xmin=0, xmax=#1,
            width=\linewidth,
            height=0.62\linewidth,
            grid=major,
            legend pos=south west,
            legend style={font=\tiny, draw=black, fill=white,
                          legend columns=3, /tikz/every even column/.append style={column sep=4pt}},
            legend cell align=left,
        ]
            \addplot[draw=none, name path=upper, forget plot]
                table[x=round, y=p90_score, col sep=comma] {#2};
            \addplot[draw=none, name path=lower, forget plot]
                table[x=round, y=p10_score, col sep=comma] {#2};
            \addplot[fill=blue!10, forget plot] fill between[of=upper and lower];
            \addplot[teal, semithick, densely dashed, mark=none]
                table[x=round, y=p90_score, col sep=comma] {#2};
            \addlegendentry{$P_{90}$}
            \addplot[orange, semithick, densely dashed, mark=none]
                table[x=round, y=p10_score, col sep=comma] {#2};
            \addlegendentry{$P_{10}$}
            \addplot[blue, very thick, mark=none]
                table[x=round, y=mean_score, col sep=comma] {#2};
            \addlegendentry{Mean}
        \end{axis}
    \end{tikzpicture}%
}

\begin{figure}[b]
    \centering
    \scoreGraph{59}{appendix/data/femnist_scores.csv}
    \caption{$\spadesuit$-KD client-recorded scores over the course of collaborative training in the \textit{FEMNIST} scenario.}
    \label{fig:scores-femnist}
\end{figure}
\begin{figure}
    \centering
    \scoreGraph{39}{appendix/data/mnist_scores.csv}
    \caption{$\spadesuit$-KD client-recorded scores over the course of collaborative training in the \textit{MNIST} scenario.}
    \label{fig:scores-mnist}
\end{figure}
\begin{figure}
    \centering
    \scoreGraph{59}{appendix/data/speech_scores.csv}
    \caption{$\spadesuit$-KD client-recorded scores over the course of collaborative training in the \textit{SpeechCommands} scenario.}
    \label{fig:scores-speech}
\end{figure}

%% file: appendix/build-graphs.tex
\renewcommand{\progRounds}{}
\renewcommand{\progMetric}{}

\renewcommand{\subfigWidth}{0.32\textwidth}

\newif\iffedkd
\fedkdtrue

\renewcommand{\lossPanel}[3]{%
\begin{subfigure}[b]{\subfigWidth}
    \centering
    \begin{tikzpicture}
        \begin{axis}[
            font=\scriptsize,
            ylabel={\textsc{\progMetric}},
            xlabel={\textsc{rounds}},
            xlabel near ticks,
            xmin=1, xmax=\progRounds,
            width=\linewidth,
            height=0.8\linewidth,
            grid=major,
        ]
            \iffedkd
            \addplot[color=gray!80!black, mark=none]        table [x expr=\coordindex + 1, y=FedKD,              col sep=comma, header=true] {#1};
            \fi
            \addplot[color=blue, mark=none]                 table [x expr=\coordindex + 1, y=DecentralizedKD,    col sep=comma, header=true] {#1};
            \addplot[color=teal, mark=none]                 table [x expr=\coordindex + 1, y=DecentralizedKD_PS, col sep=comma, header=true] {#1};
            \addplot[color=violet, mark=none]               table [x expr=\coordindex + 1, y=DecentralizedA2KD,  col sep=comma, header=true] {#1};
            \addplot[color=orange, mark=none, line width=1.1pt] table [x expr=\coordindex + 1, y=SpadeKD,         col sep=comma, header=true] {#1};
        \end{axis}
    \end{tikzpicture}
    \caption{#2}
    \label{#3}
\end{subfigure}%
}

\renewcommand{\legendItem}[2]{%
    \tikz[baseline=-0.5ex]{\draw[#1] (0,0) -- (0.45,0);}~#2}
\renewcommand{\methodsLegend}{{\scriptsize
    \iffedkd\legendItem{gray!80!black, line width=1pt}{FedKD}\quad\fi
    \legendItem{blue,          line width=1pt}{D-KD}\quad
    \legendItem{teal,          line width=1pt}{D-KD-PS}\quad
    \legendItem{violet,        line width=1pt}{D-AKD}\quad
    \legendItem{orange,        line width=1.1pt}{SPADE-KD}%
}}

\renewcommand{\buildGraph}[9]
{%
\renewcommand{\progRounds}{#1}
\renewcommand{\progMetric}{#2}
\lossPanel{#3_mean.csv}{#4}{#5}\hfill
\lossPanel{#3_p10.csv}{#6}{#7}\hfill
\lossPanel{#3_p90.csv}{#8}{#9}%
}

\fedkdtrue

\begin{figure*}
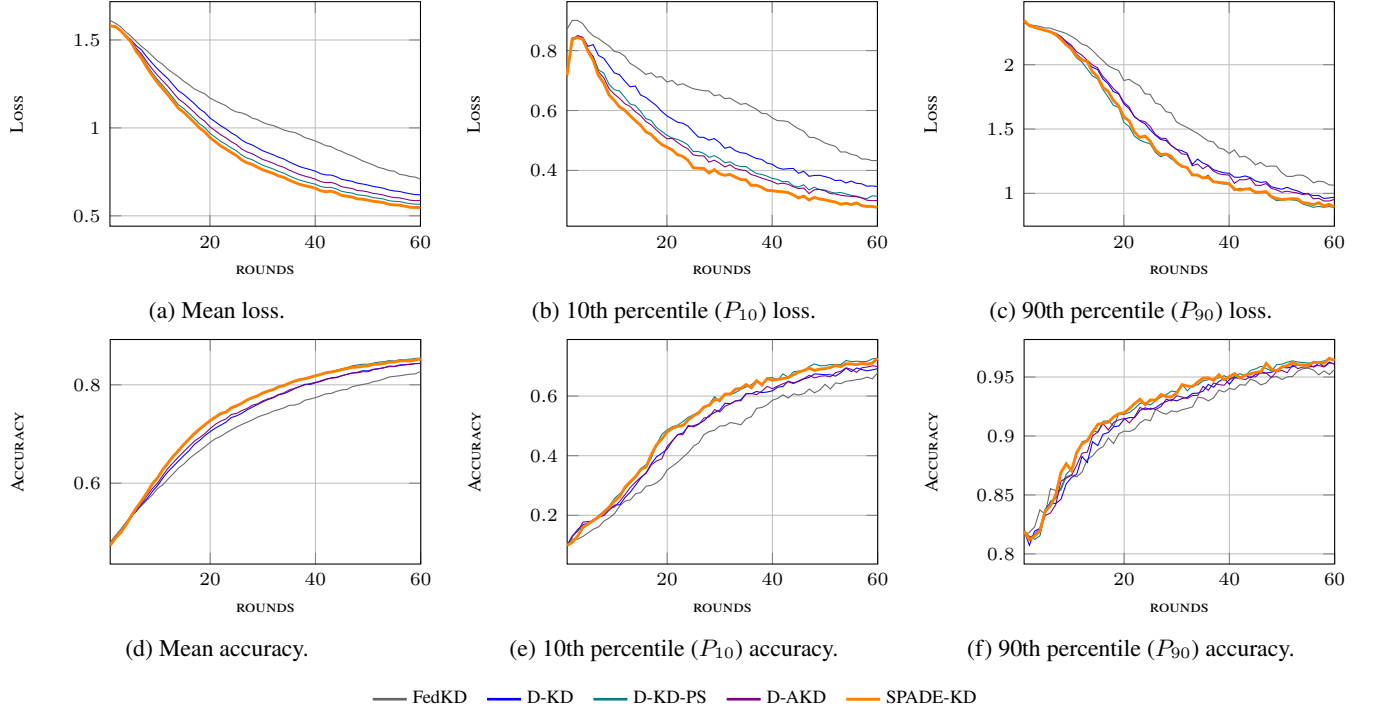

    \centering
    \buildGraph{60}{Loss}{plots/data/femnist_loss}
        {Mean loss.}{res:app-femnist-loss-mean}
        {10th percentile ($P_{10}$) loss.}{res:app-femnist-loss-p10}
        {90th percentile ($P_{90}$) loss.}{res:app-femnist-loss-p90}
    \\[0.4em]
    \buildGraph{60}{Accuracy}{plots/data/femnist_accuracy}
        {Mean accuracy.}{res:app-femnist-acc-mean}
        {10th percentile ($P_{10}$) accuracy.}{res:app-femnist-acc-p10}
        {90th percentile ($P_{90}$) accuracy.}{res:app-femnist-acc-p90}
    \\[0.4em]
    \methodsLegend
    \caption{Clients' loss and accuracy on the \textit{FEMNIST} scenario (naturally heterogeneous image data).}
    \label{res:app-femnist}
\end{figure*}

\begin{figure*}
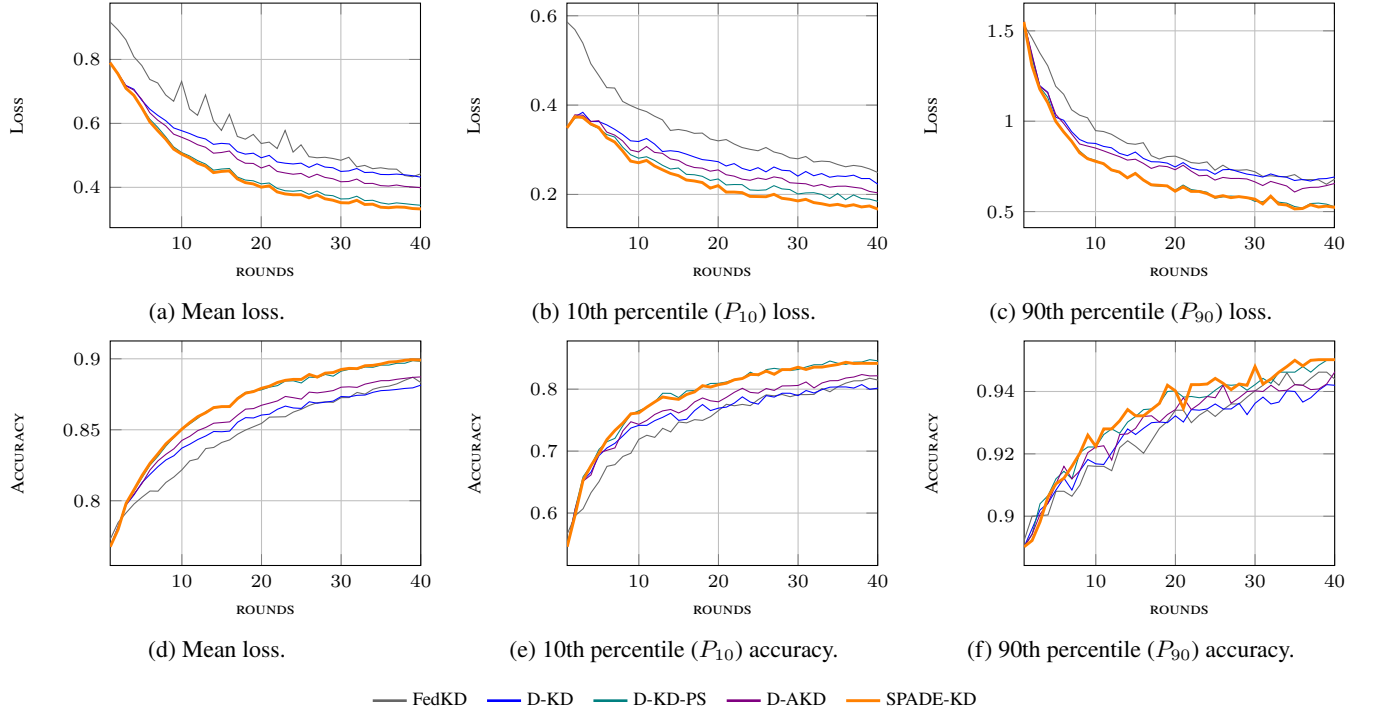

    \centering
    \buildGraph{40}{Loss}{plots/data/mnist_loss}
        {Mean loss.}{res:app-mnist-loss-mean}
        {10th percentile ($P_{10}$) loss.}{res:app-mnist-loss-p10}
        {90th percentile ($P_{90}$) loss.}{res:app-mnist-loss-p90}
    \\[0.4em]
    \buildGraph{40}{Accuracy}{plots/data/mnist_accuracy}
        {Mean loss.}{res:app-mnist-acc-mean}
        {10th percentile ($P_{10}$) accuracy.}{res:app-mnist-acc-p10}
        {90th percentile ($P_{90}$) accuracy.}{res:app-mnist-acc-p90}
    \\[0.4em]
    \methodsLegend
    \caption{Clients' loss and accuracy on the \textit{MNIST} scenario (artificially heterogeneous image data).}
    \label{res:app-mnist}
\end{figure*}

\begin{figure*}
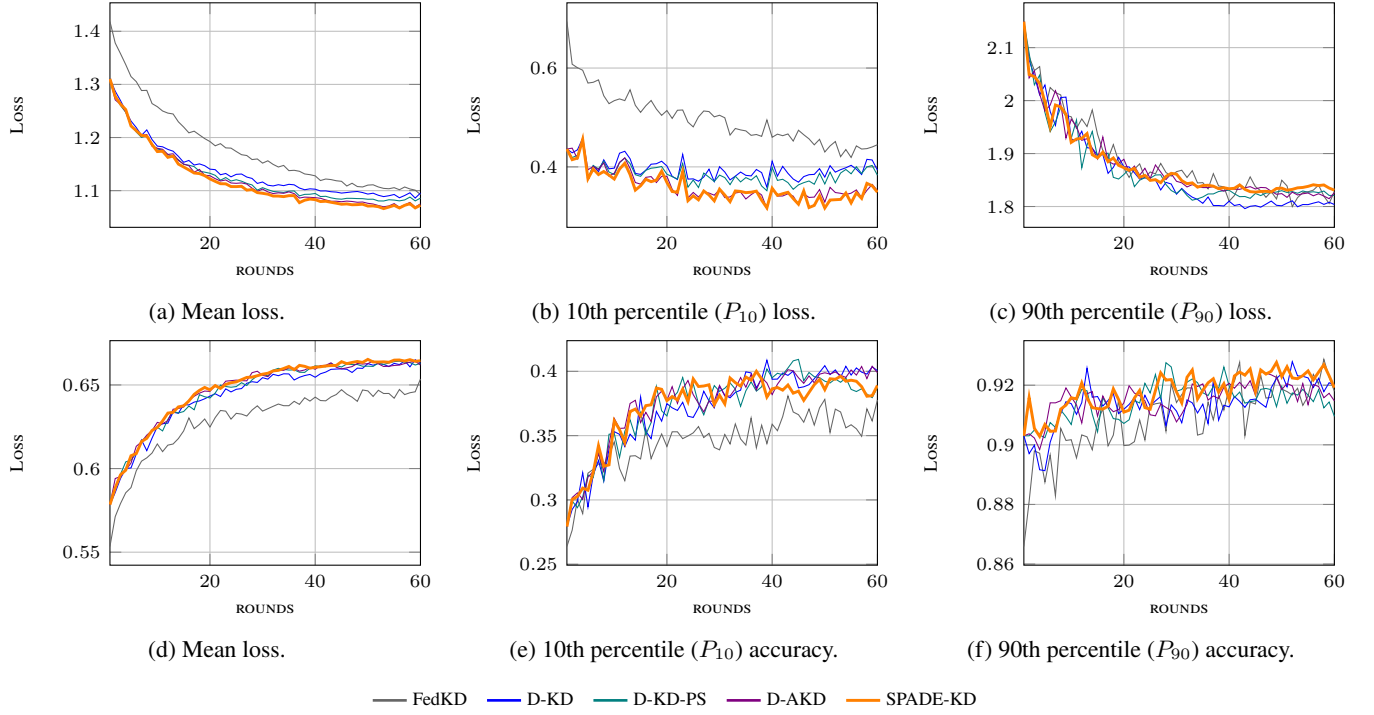

    \centering
    \buildGraph{60}{Loss}{plots/data/speech_loss}
        {Mean loss.}{res:app-speech-loss-mean}
        {10th percentile ($P_{10}$) loss.}{res:app-speech-loss-p10}
        {90th percentile ($P_{90}$) loss.}{res:app-speech-loss-p90}
    \\[0.4em]
    \buildGraph{60}{Loss}{plots/data/speech_accuracy}
        {Mean loss.}{res:app-speech-acc-mean}
        {10th percentile ($P_{10}$) accuracy.}{res:app-speech-acc-p10}
        {90th percentile ($P_{90}$) accuracy.}{res:app-speech-acc-p90}
    \\[0.4em]
    \methodsLegend
    \caption{Clients' loss and accuracy on the \textit{SpeechCommands} scenario (naturally heterogeneous audio data).}
    \label{res:app-speech}
\end{figure*}

%% file: appendix/build-comparison-table.tex
%
%
%
%

\begin{filecontents*}{appendix/data/femnist_loss_comparison.csv}
scenario,method,column,round,mean_loss,mean_impr_pct,p10_loss,p10_impr_pct,p90_loss,p90_impr_pct
femnist,D-KD,DecentralizedKD,60,0.619609,0.0,0.346189,0.0,0.969468,0.0
femnist,D-KD-PS,DecentralizedKD_PS,60,0.565978,8.66,0.313705,9.38,0.89031,8.17
femnist,P2P-AKD,P2P_AKD,60,0.529401,14.56,0.252592,27.04,0.905281,6.62
femnist,SPADE-KD,SpadeKD,60,0.54755,11.63,0.276332,20.18,0.894166,7.77
femnist,FedKD,FedKD,60,0.70837,-14.33,0.433263,-25.15,1.063083,-9.66
femnist,D-AKD,DecentralizedA2KD,60,0.586472,5.35,0.299518,13.48,0.95276,1.72
femnist,D-AKD-old,DecentralizedAKD,60,0.548599,11.46,0.253959,26.64,0.938683,3.18
\end{filecontents*}
\DTLloaddb{cmpLossFem}{appendix/data/femnist_loss_comparison.csv}
\begin{filecontents*}{appendix/data/mnist_loss_comparison.csv}
scenario,method,column,round,mean_loss,mean_impr_pct,p10_loss,p10_impr_pct,p90_loss,p90_impr_pct
mnist,D-KD-PS,DecentralizedKD_PS,40,0.344046,20.36,0.18475,17.36,0.526901,23.78
mnist,P2P-AKD,P2P_AKD,40,0.331644,23.23,0.160995,27.98,0.525015,24.05
mnist,FedKD,FedKD,40,0.442065,-2.33,0.24946,-11.59,0.67867,1.82
mnist,D-AKD,DecentralizedA2KD,40,0.398958,7.65,0.203625,8.91,0.655142,5.22
mnist,D-KD,DecentralizedKD,40,0.432007,0.0,0.223547,0.0,0.69125,0.0
mnist,SPADE-KD,SpadeKD,40,0.33256,23.02,0.167571,25.04,0.523576,24.26
mnist,D-AKD-old,DecentralizedAKD,40,0.370243,14.3,0.172506,22.83,0.621692,10.06
\end{filecontents*}
\DTLloaddb{cmpLossMni}{appendix/data/mnist_loss_comparison.csv}
\begin{filecontents*}{appendix/data/speech_loss_comparison.csv}
scenario,method,column,round,mean_loss,mean_impr_pct,p10_loss,p10_impr_pct,p90_loss,p90_impr_pct
speech,D-KD-PS,DecentralizedKD_PS,60,1.085947,0.87,0.383181,2.61,1.820869,-0.94
speech,D-AKD,DecentralizedA2KD,60,1.075052,1.86,0.352128,10.5,1.826152,-1.24
speech,D-KD,DecentralizedKD,60,1.09545,0.0,0.393436,0.0,1.803849,0.0
speech,FedKD,FedKD,60,1.09256,0.26,0.444811,-13.06,1.828018,-1.34
speech,SPADE-KD,SpadeKD,60,1.072557,2.09,0.348643,11.39,1.830869,-1.5
\end{filecontents*}
\DTLloaddb{cmpLossSpe}{appendix/data/speech_loss_comparison.csv}
\begin{filecontents*}{appendix/data/femnist_accuracy_comparison.csv}
scenario,method,column,round,mean_accuracy,mean_impr_pct,p10_accuracy,p10_impr_pct,p90_accuracy,p90_impr_pct
femnist,D-KD,DecentralizedKD,60,0.843545,0.0,0.691579,0.0,0.96078,0.0
femnist,D-KD-PS,DecentralizedKD_PS,60,0.854036,1.24,0.726638,5.07,0.963755,0.31
femnist,P2P-AKD,P2P_AKD,60,0.845825,0.27,0.710905,2.79,0.960399,-0.04
femnist,SPADE-KD,SpadeKD,60,0.852351,1.04,0.726845,5.1,0.963988,0.33
femnist,FedKD,FedKD,60,0.827565,-1.89,0.678299,-1.92,0.956397,-0.46
femnist,D-AKD,DecentralizedA2KD,60,0.84307,-0.06,0.69925,1.11,0.961053,0.03
femnist,D-AKD-old,DecentralizedAKD,60,0.840852,-0.32,0.699317,1.12,0.957173,-0.38
\end{filecontents*}
\DTLloaddb{cmpAccFem}{appendix/data/femnist_accuracy_comparison.csv}
\begin{filecontents*}{appendix/data/mnist_accuracy_comparison.csv}
scenario,method,column,round,mean_accuracy,mean_impr_pct,p10_accuracy,p10_impr_pct,p90_accuracy,p90_impr_pct
mnist,D-KD-PS,DecentralizedKD_PS,40,0.89822,1.91,0.8458,5.51,0.9502,0.87
mnist,P2P-AKD,P2P_AKD,40,0.89954,2.06,0.8436,5.24,0.952,1.06
mnist,FedKD,FedKD,40,0.8833,0.21,0.8148,1.65,0.944,0.21
mnist,D-AKD,DecentralizedA2KD,40,0.88724,0.66,0.8216,2.5,0.9462,0.45
mnist,D-KD,DecentralizedKD,40,0.88142,0.0,0.8016,0.0,0.942,0.0
mnist,SPADE-KD,SpadeKD,40,0.89908,2.0,0.8416,4.99,0.9502,0.87
mnist,D-AKD-old,DecentralizedAKD,40,0.88946,0.91,0.825,2.92,0.9502,0.87
\end{filecontents*}
\DTLloaddb{cmpAccMni}{appendix/data/mnist_accuracy_comparison.csv}
\begin{filecontents*}{appendix/data/speech_accuracy_comparison.csv}
scenario,method,column,round,mean_accuracy,mean_impr_pct,p10_accuracy,p10_impr_pct,p90_accuracy,p90_impr_pct
speech,D-KD-PS,DecentralizedKD_PS,60,0.662868,-0.02,0.387371,-3.29,0.909554,-1.15
speech,D-AKD,DecentralizedA2KD,60,0.661906,-0.17,0.398971,-0.4,0.914731,-0.58
speech,D-KD,DecentralizedKD,60,0.663025,0.0,0.400554,0.0,0.92011,0.0
speech,FedKD,FedKD,60,0.653583,-1.42,0.377754,-5.69,0.917962,-0.23
speech,SPADE-KD,SpadeKD,60,0.664498,0.22,0.389031,-2.88,0.919178,-0.1
\end{filecontents*}
\DTLloaddb{cmpAccSpe}{appendix/data/speech_accuracy_comparison.csv}

\newcommand{\cmpval}[4]{
  \begingroup\edef\raw{\csname V#1#2#3#4\endcsname}%
  \pgfmathprintnumber[fixed,fixed zerofill,precision=3]{\raw}\endgroup}
\renewcommand{\cmppct}[4]{
  \begingroup\edef\raw{\csname P#1#2#3#4\endcsname}\def\midarg{#3}\def\dkdc{DKD}%
  \ifx\midarg\dkdc$-$\else\textit{\pgfmathprintnumber[fixed,precision=1,print sign,fixed zerofill]{\raw}}\fi
  \endgroup}
\renewcommand{\cmprow}[3]{
  #2
  & \cmpval{#1}{Fem}{#3}{mean} & \cmppct{#1}{Fem}{#3}{mean} & \cmpval{#1}{Fem}{#3}{p10} & \cmppct{#1}{Fem}{#3}{p10} & \cmpval{#1}{Fem}{#3}{p90} & \cmppct{#1}{Fem}{#3}{p90}
  & \cmpval{#1}{Mni}{#3}{mean} & \cmppct{#1}{Mni}{#3}{mean} & \cmpval{#1}{Mni}{#3}{p10} & \cmppct{#1}{Mni}{#3}{p10} & \cmpval{#1}{Mni}{#3}{p90} & \cmppct{#1}{Mni}{#3}{p90}
  & \cmpval{#1}{Spe}{#3}{mean} & \cmppct{#1}{Spe}{#3}{mean} & \cmpval{#1}{Spe}{#3}{p10} & \cmppct{#1}{Spe}{#3}{p10} & \cmpval{#1}{Spe}{#3}{p90} & \cmppct{#1}{Spe}{#3}{p90} \\}

\newcommand{\cmpfetch}[3]{
  \foreach \sc in {Fem,Mni,Spe}{%
    \edef\dbname{cmp#2\sc}%
    \foreach \m/\mid in {FedKD/FEDKD, D-KD/DKD, D-KD-PS/DKDPS, D-AKD/DAKD, SPADE-KD/SPADE}{%
      \foreach \st in {mean,p10,p90}{%
        \DTLgetvalueforkey{\tmp}{\st_#3}{\dbname}{method}{\m}\expandafter\xdef\csname V#1\sc\mid\st\endcsname{\tmp}%
        \DTLgetvalueforkey{\tmp}{\st_impr_pct}{\dbname}{method}{\m}\expandafter\xdef\csname P#1\sc\mid\st\endcsname{\tmp}%
      }%
    }%
  }%
}
\newcommand{\maketable}[6]{
  \cmpfetch{#1}{#2}{#3}%
  \begin{table*}[]
    \centering
    \caption{#5}
    \label{#6}
    \footnotesize
    \setlength{\tabcolsep}{2pt}
    \begin{tabular}{l|rrrrrr|rrrrrr|rrrrrr}
      \hline
      & \multicolumn{6}{c|}{\textit{FEMNIST}}
      & \multicolumn{6}{c|}{\textit{MNIST}}
      & \multicolumn{6}{c}{\textit{SpeechCommands}}\\
      & \multicolumn{2}{c}{Mean} & \multicolumn{2}{c}{$P_{10}$} & \multicolumn{2}{c|}{$P_{90}$}
      & \multicolumn{2}{c}{Mean} & \multicolumn{2}{c}{$P_{10}$} & \multicolumn{2}{c|}{$P_{90}$}
      & \multicolumn{2}{c}{Mean} & \multicolumn{2}{c}{$P_{10}$} & \multicolumn{2}{c}{$P_{90}$}\\
      Method & #4 & $\Delta\%$ & #4 & $\Delta\%$ & #4 & $\Delta\%$ & #4 & $\Delta\%$ & #4 & $\Delta\%$ & #4 & $\Delta\%$ & #4 & $\Delta\%$ & #4 & $\Delta\%$ & #4 & $\Delta\%$\\
      \hline
      \cmprow{#1}{FedKD}{FEDKD}
      \cmprow{#1}{D-KD}{DKD}
      \cmprow{#1}{D-KD-PS}{DKDPS}
      \cmprow{#1}{D-AKD}{DAKD}
      \cmprow{#1}{\textbf{$\spadesuit$-KD}}{SPADE}
      \hline
    \end{tabular}
  \end{table*}
}

\maketable{L}{Loss}{loss}{L}%
  {Final-round client \emph{loss} (lower is better) and percentage improvement $\Delta\%$ versus the naive decentralized KD (D-KD) baseline, at the mean, $P_{10}$, and $P_{90}$ across clients. $\Delta\%>0$ means lower loss than D-KD.}%
  {tab:app-comparison-loss}

\maketable{A}{Acc}{accuracy}{Acc.}%
  {Final-round client \emph{accuracy} (higher is better) and percentage improvement $\Delta\%$ versus the naive decentralized KD (D-KD) baseline, at the mean, $P_{10}$, and $P_{90}$ across clients. $\Delta\%>0$ means higher accuracy than D-KD.}%
  {tab:app-comparison-acc}